# Super ensemble learning for daily streamflow forecasting: Large-scale demonstration and comparison with multiple machine learning algorithms

Hristos Tyralis[1,2], Georgia Papacharalampous[3], and Andreas Langousis[4]

[1]Department of Water Resources and Environmental Engineering, School of Civil Engineering, National Technical University of Athens, Iroon Polytechniou 5, 157 80 Zografou, Greece

[2]Air Force Support Command, Hellenic Air Force, Elefsina Air Base, 192 00 Elefsina, Greece (https://orcid.org/0000-0002-8932-4997)

[3]Department of Water Resources and Environmental Engineering, School of Civil Engineering, National Technical University of Athens, Iroon Polytechniou 5, 157 80 Zografou, Greece (https://orcid.org/0000-0001-5446-954X), (papacharalampous.georgia@gmail.com)

[4]Department of Civil Engineering, School of Engineering, University of Patras, University Campus, Rio, 26 504, Patras, Greece (https://orcid.org/0000-0002-0643-2520), (andlag@alum.mit.edu)

Corresponding author: Hristos Tyralis (hristos@itia.ntua.gr, montchrister@gmail.com)

**Abstract**: Daily streamflow forecasting through data-driven approaches is traditionally performed using a single machine learning algorithm. Existing applications are mostly restricted to examination of few case studies, not allowing accurate assessment of the predictive performance of the algorithms involved. Here we propose super learning (a type of ensemble learning) by combining 10 machine learning algorithms. We apply the proposed algorithm in one-step ahead forecasting mode. For the application, we exploit a big dataset consisting of 10-year long time series of daily streamflow, precipitation and temperature from 511 basins. The super learner improves over the performance of the linear regression algorithm by 20.06%, outperforming the "hard to beat in practice" equal weight combiner. The latter improves over the performance of the linear regression algorithm by 19.21%. The best performing individual machine learning algorithm is neural networks, which improves over the performance of the linear regression algorithm by 16.73%, followed by extremely randomized trees (16.40%), XGBoost (15.92%), loess (15.36%), random forests (12.75%), polyMARS (12.36%), MARS (4.74%), lasso (0.11%) and support vector regression (−0.45%). Based on the obtained large-scale results, we propose super learning for daily streamflow forecasting.

**Keywords**: combining forecasts; ensemble learning; hydrology; stacking

## 1. Introduction

Streamflow forecasting at various temporal scales and time steps ahead is important for engineering purposes (e.g. hydro-power generation, dam regulation and other water resources engineering purposes), as well as environmental and societal purposes (e.g. flood protection and long-term water resources planning). Here, we are interested in one-step ahead daily streamflow forecasting.

In streamflow forecasting, the predictive ability of the implemented model is of high importance; therefore, more flexible albeit less interpretable models (e.g. machine learning algorithms) are acceptable, given that they are more accurate. While accuracy is important in engineering, the current trend in the field of hydrology favours model interpretability (see e.g. [11]). The reader is referred to [16], [36, pp 24–26)] and [60], for a general discussion on the issue of interpretability versus flexibility or, equivalently, understanding versus prediction in algorithmic modelling. Here focus is on accuracy.

The dominant approach in daily streamflow forecasting is the implementation of machine learning regression algorithms, while linear models (mostly time series models) have been found to be more competitive at larger time scales (e.g. monthly and annual; [49,50]). Regression algorithms model the dependent variable (streamflow at some time) as function of a set of selected predictor variables (e.g. past streamflow values, past precipitation values, and past temperature values). In the case of machine learning regression, this function is learnt directly from data through an algorithmic approach. Popular algorithms include neural networks (see e.g. [1,21,41,65]), support vector machines [56], decision trees, random forests and their variants [72], with numerous algorithmic variants (see e.g. [23] for the most representative ones) having been more or less applied to hydrologic case studies. Note, however, that existing approaches to daily streamflow forecasting are mostly based on the implementation of a single machine learning algorithm.

Combining forecasts from different methods has been proved to increase the forecasting accuracy. This point was initially raised by [7], while the argumentation in favour of forecast combinations, referred to as *"ensemble learning"* in the literature, was further strengthened in the early 90's (see e.g. [32,80]). The "*no free lunch theorem*" [89] implies that no universally best machine learning algorithm exists. Thus, ensemble learning, i.e. combining multiple machine learning algorithms (hereinafter termed as base-learners) instead of using a single one, may increase the predictive accuracy of the forecasts. Overviews of model combinations in general and ensemble learning in particular can be found in [24] and [59], respectively. Here we are interested in stacked generalization (also referred to as stacking), a particular type of ensemble learning where base-learners are properly weighted so certain performance metrics are minimized (see e.g. [53,74] for specific applications in probabilistic hydrological post-processing), which was initially suggested by [88] and later investigated by [14] for regression.

The simplest combination of models is equal weight averaging. The latter combination approach has been proved *"hard to beat in practice"* by more complex combination methods, a finding that has been termed "*forecast combination puzzle*" by [66]. While research on the causes of the "*forecast combination puzzle*" is currently inconclusive (see e.g. [19,63,70]), one can intuitively attribute its sources to the fact that as the level uncertainty (or equivalently the number of base-learners) increases, weight optimization may not lead to significant improvements relative to simple averaging; i.e. a uniform weighting scheme that assigns equal weights to all base-learners (see [74]).

Most published studies focusing on daily streamflow forecasting use small datasets (e.g. data collected from a couple of rivers) to present some type of new method, usually referred to as hybrid when combining e.g. neural networks with an optimization algorithm. While such studies may be useful from a hydrological standpoint, the obtained results cannot be conclusive regarding the accuracy of the proposed method, due to the high degree of randomness induced by sample variability. While small-scale applications were acceptable in the early era of neural network hydrology, the current status of data availability allows for large-scale applications. Actually, recent studies based on big datasets have revealed ground breaking results in the field of hydrological forecasting (see e.g. [49,52]), as large-scale applications allow for less biased simulation designs to assess the relative performance of new and existing methods (see e.g. the commentary in

[12]).

The aim of our study is to propose a new method for streamflow forecasting based on a stacking algorithm, specifically super learning [77]. Along with the introduction of the new approach, our study aims at advancing the existing knowledge and current state-of-the-art in the field of machine learning by:

a. Introducing a super learning framework to combine 10 machine learning algorithms, and comparing super learning with the "*hard to beat in practice*" equal weight combiner.

b. Assessing the relative performance of 10 individual machine learning algorithms in daily streamflow forecasting.

c. Using more than 500 streamflow time series to support the quantitative conclusions reached.

Beyond presentation of the new method (see remark (a) above), we consider remarks (b) and (c) equally important, since most studies in the field use small datasets (i.e. formed by a single-digit number of time series) to compare a limited number of machine learning algorithms. Use of big datasets can provide insights and facilitate understanding and contrasting of the properties of various algorithms in predicting daily streamflow, consisting an important asset for engineering applications.

## 2. Methods

In this Section, we present short descriptions of the individual machine learning algorithms (base-learners) used (please note that an exhaustive presentation of the algorithms is out of the scope of the present study), the three combiner learners (i.e. super learner, equal weight combiner, and best learner), as well as the variable selection methodology, the metrics used to assess the relative performance of the algorithms, and the testing procedure.

### 2.1 Base-learners

A detailed description of the majority of the base-learners exploited herein is out of the scope of the manuscript and can be found in [33,36]. All algorithms have been implemented and documented in the R programming language. Details on their software implementation can be found in Appendix A. To ensure reproducibility of the results, Appendix A also includes the versions of the software packages used herein.

### 2.1.1 Linear regression

Linear regression is the simplest model used herein. It is described in detail by [33 pp 43–55]. The dependent variable is modelled as a linear combination of the predictor variables, while the weights are estimated by minimizing the residual sum of squares (least squares method).

### 2.1.2 Lasso

The least absolute shrinkage and selection operator (lasso) algorithm [69] performs variable selection and regularization by imposing the lasso penalty ($L_1$ shrinkage) in the least squares method, aiming to shrink its coefficients, while allowing for elimination of non-influential predictor variables by nullifying their coefficients.

### 2.1.3 Loess

Locally estimated scatterplot smoothing (loess, [20]) fits a polynomial surface (determined by the predictor variables) to the data by using local fitting. Here we used a second-degree polynomial.

### 2.1.4 Multivariate adaptive regression splines

Multivariate adaptive regression splines (MARS, [25,26]) is a weighted sum of basis functions, with total number and associated parameters (i.e. product degree and knot locations, respectively) being automatically determined from data. Here we build an additive model (i.e. a model without interactions), where the predictor variables enter the regression through a linear sum of hinge basis functions.

### 2.1.5 Multivariate adaptive polynomial spline regression

Multivariate adaptive polynomial spline regression (polyMARS, [40,67]) is an adaptive regression procedure that uses piecewise linear splines to model the dependent variable. It is similar to MARS, with main differences being that "*(a) it requires linear terms of a predictor to be in the model before nonlinear terms using the same predictor can be added and (b) it requires a univariate basis function to be in the model before a tensor-product basis function involving the univariate basis function can be in the model*" [39].

### 2.1.6 Random forests

Random forests [15] are bagging (abbreviation for bootstrap aggregation) of regression trees with an additional degree of randomization; i.e. they randomly select a fixed number of predictor variables as candidates when determining the nodes of the decision tree.

*2.1.7 XGBoost*

Extreme Gradient Boosting (XGBoost, [17]) is an implementation of gradient boosted decision trees (see e.g. [27,43,45], albeit considerably faster and better performing. Gradient boosting is an approach that creates new models (in this case decision trees) to predict the errors of prior models. The final model adds all fitted models. A gradient descent algorithm is used to minimize the loss function when adding new decision trees. XGBoost uses a model formalization that is more regularized to control over-fitting. This procedure renders XGBoost more accurate than gradient boosting.

*2.1.8 Extremely randomized trees*

Extremely randomized trees [31] are similar to random forests. These two models mostly differ in the splitting procedure. Contrary to random forests, in extremely randomized trees the cut-point is fully random.

*2.1.9 Support vector machines*

The principal concept of support vector regression is to estimate a linear regression model in a high dimensional feature space. In this space the input data are mapped using a (nonlinear) kernel function [64,78]. Here we used a radial basis kernel.

*2.1.10 Neural networks*

The principal concept of neural networks is to extract linear combinations of the predictor variables as derived features, and then model the dependent variable as a nonlinear function of these features [33, p 389]. Here we used feed-forward neural networks [57, pp 143–180].

## 2.2 Super learning

Super learner is a convex weighted combination of multiple machine learning algorithms, with weights that sum to unity and are equal or higher than zero (see [75,76,77]). The weights are estimated through a *k*-fold cross-validation procedure (here we choose *k* = 5) in the training set (see Section 2.5), so that a properly selected loss function is minimized. Here we minimize the quadratic loss function, which is equivalent to minimizing the root mean squared error (RMSE). Then the base-learners are retrained in the full training dataset, and the super learner predictions are obtained as the weighted sum (using the estimated weights of the cross-validation procedure) of the retrained base-learners.

Super learning (as every stacking algorithm) can combine ensemble learners (e.g. bagging algorithms, boosting algorithms and more) and different types of base learners, while e.g. bagging or boosting algorithms, use a single type of base learners.

## 2.3 Other ensemble learners

In addition to super learning, we applied the equal weight combiner by assigning a uniform weighting scheme (i.e. weights equal to 1/10) to all base-learners. Furthermore, we used an ensemble learner (referred to as best learner), which selects the best base-learner based on its performance in the *k*-fold cross-validation procedure in the training set.

## 2.4 Variable selection

Variable selection constitutes a complex problem that has been extensively investigated with no exact solution (see e.g. [34]) as selection of variables is strictly linked to the problem at hand. In daily streamflow forecasting, daily streamflow $q_i$ may depend on past streamflow, precipitation and temperature values ($q_j$, $p_j$, $t_j$, $j = 1, ..., i – 1$, respectively). If non-informative predictor variables are included in the model, the performance of some algorithms (e.g. linear regression) may decrease considerably, while if too many predictor variables are included, the computational burden may become prohibitive. Missing some informative predictor variables may also harm the performance of the model.

Several strategies can be employed to select predictor variables, e.g. an exhaustive search [71], use of correlation measures, partial mutual information [42] and the like. An overview of variable selection procedures in water resources engineering can be found in [13].

Here we selected to use the permutation variable importance metric (VIM) used by random forests algorithms for variable selection. The permutation VIM measures the mean decrease in accuracy in the out-of-bag (OOB) sample by randomly permuting the predictor variable of interest. OOB samples are the samples remaining after bootstrapping the training set (see also Section 2.1.6). VIM permits ranking of the relative significance of predictor variables [72] and is a commonly used variable selection procedure. We computed VIM of daily values of streamflow, precipitation and temperature of the last month, i.e. 90 predictor variables in total. We selected the resulting five most important predictor variables for each process type (i.e., streamflow, precipitation and temperature). The fitting problem is formulated as:

$$q_i = f(\{q_j, p_k, t_l\}), j, k, l \in \{\text{five values in } (i-30, \ldots, i-1)\} \quad (1)$$

If some of the best ranked possible predictor variables display negative VIM values, they are excluded from the set of predictor variables, since they are non informative (see e.g. [73] and references therein). In this case, the set of predictor variables includes less than 15 variables.

## 2.5 Training and testing

Machine learning algorithms in regression settings approximate the function $f$ in eq. (1) through training on data. During training, some hyperparameter optimization can be performed to enhance the performance of the model. However, default hyperparameter values used in software implementations usually display favourable properties, as e.g. proved in large scale empirical studies in hydrology [51], while hyperparameter optimization may be computationally costly with little improvement in performance. That said, in the present study, we decided to use default hyperparameter values, as suggested in the corresponding software implementations (see Appendix A).

To estimate the generalization error of the implemented algorithms, one should compare to an independent set; i.e. a set not used for training, termed as test set. Following recent theoretical studies [4], we use equally sized training and test sets to assess the performance of the algorithms. Following relevant suggestions by [6], we do not use hypothesis tests to report the significance of the differences, as their use in the field of forecasting may lead to misinterpretations. Instead, we prefer to use "*effect sizes*", as e.g. done in forecasting competitions [6], which in our case are "*percent error reductions*" in terms of a specified metric. Our choice also overcomes the problems of (a) computing significance of the forecasting performance differences between every couple of the implemented algorithms and (b) using some type of scaled metrics (e.g. the widely used in hydrology Nash –Suttcliffe efficiency) which is usually accompanied by other disadvantages. Please also note that relative improvements between the benchmark and the examined algorithm are equal for the cases of using RMSE (see for definition Section 2.6) and the Nash –Suttcliffe efficiency.

## 2.6 Metrics

Although the super learner is optimized with respect to RMSE, we use multiple metrics to understand the effect of this optimization and quantitatively assess the relative performance of the algorithms. An overview of metrics that can be used to assess the performance of forecasting methods can be found in [35]. Here we use RMSE, the mean of absolute errors (MAE), the median of absolute errors (MEDAE) and the squared correlation $r^2$ between the forecasts $f_n$ and the observations $o_n$. All metrics, defined by the following equations, are computed in the testing period.

$$E_n := f_n - o_n \tag{2}$$

$$\text{MAE} := (1/|N|) \sum_n |E_n| \tag{3}$$

$$\text{RMSE} := ((1/|N|) \sum_n E_n^2)^{1/2} \tag{4}$$

$$\text{MEDAE} := \text{median}_n\{|E_n|\} \tag{5}$$

$$r^2 := (\text{corr}(\boldsymbol{f}, \boldsymbol{o}))^2 \tag{6}$$

In Equation (6), $\boldsymbol{f}$ and $\boldsymbol{o}$ denote the vectors of the forecasts and observations, respectively in the testing period. MAE, RMSE and MEDAE take values in the range [0, ∞), with 0 values indicating perfect forecasts. $r^2$ takes values in the range [0, 1], with values equal to 1 denoting perfect forecasts.

## 3. Data and application

### 3.1 Data

We used the Catchment Attributes and MEteorology for Large-sample Studies (CAMELS) dataset; for details see [2,3,46,47,48,68]. The dataset includes daily minimum temperatures, maximum temperatures, precipitation and streamflow data from 671 small- to medium-sized basins in the contiguous US (CONUS). The mean daily temperature was estimated by averaging the minimum and maximum daily temperatures. Changes in the basins due to human influences are minimal. Here we focus on the 10-year period 2004-2013, while basins with missing data or other inconsistencies have been excluded. The final sample consists of 511 basins representing diverse climate types over CONUS; see Figure 1.

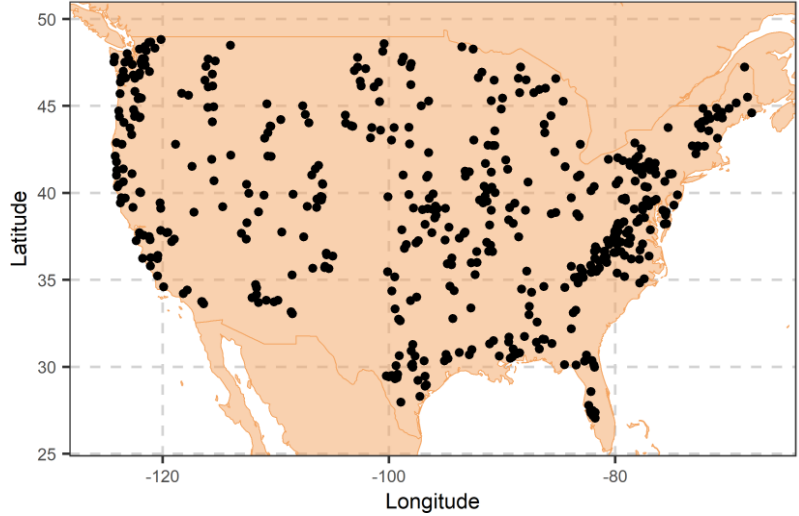

Figure 1. The 511 basins over CONUS used in the study.

## 3.2 Implementation of methods

In what follows, we detail the implementation of the algorithms and their testing.

a. The training and testing periods (hereafter denoted by $T_1$ and $T_2$, respectively) are set to $T_1$ = {2004-01-01, …, 2008-12-31} and $T_2$ = {2009-01-01, …, 2013-12-31}.

b. For an arbitrary basin, random forests VIM approach (see Section 2.4) is applied in period $T_1$ by using $q_j$, $p_k$, $t_l$, $j$, $k$, $l \in \{i - 30, ..., i - 1\}$ as predictor variables (90 predictor variables in total) and $q_i$ as dependent variable. The training sample includes 1827 instances, i.e. as many as the number of days in period $T_1$. The five most important predictor variables for each process type (i.e. $q$, $p$, $t$) are selected based on their VIM values (see Section 2.4) and used for training of the algorithms. In the case when less than five predictor variables have positive VIM values for a certain process type, then the predictor variables with negative (or zero) VIM values are excluded and the number of the selected predictor variables reduces to less than 15. The selected predictor variables are used in the next steps.

c. All algorithms of Sections 2.1.1 – 2.1.10 are trained in period $T_1$ in a 5-fold cross-validation setting.

d. The super learner (composed by the ten base-learners of step (c); see Section 2.2) is also trained in period $T_1$ using 5-fold cross-validation. This is done by estimating the 5-fold cross-validated risk for each base-learner in step (c) and computing its weight.

e. The ten trained base-learners of step (c) are retrained in the full $T_1$ period and predict streamflow in period $T_2$. The testing sample includes 1826 instances; i.e. as many as the number of days in period $T_2$.

f. The super learner (which uses the estimated weights of step (e) and weights the retrained base learners), the equal weight combiner (which averages the 10 retrained base-learners; see Section 2.3) and the best learner (i.e. the retrained base-learner with the least 5-fold cross-validated risk in period $T_1$; see Section 2.3 and step (c)) predict daily streamflow in period $T_2$.

g. The metrics of Section 2.6 are computed for each of the 13 algorithms (see steps (e) and (f)) in period $T_2$.

Finally, the metric values are summarized for all basins in period $T_2$.

## 4. Results

Here we summarize the predictive performance of the 13 algorithms in period $T_2$ for the 511 basins. We present the rankings of the algorithms (Section 4.1) and their relative improvements with respect to the linear regression benchmark (Section 4.2). An investigation on the estimated weights of the super learner is also presented (Section 4.3).

### 4.1 Ranking of methods

Figure 2 presents the mean rankings of the 13 algorithms according to their performance in terms of the examined metrics. Rankings range from 1 to 13, with lower values indicating better performance. For instance, when examining an arbitrary basin, the 13 algorithms are ranked according to their performance in terms of each metric separately. Then, these rankings are averaged over all basins, conditional on the metric.

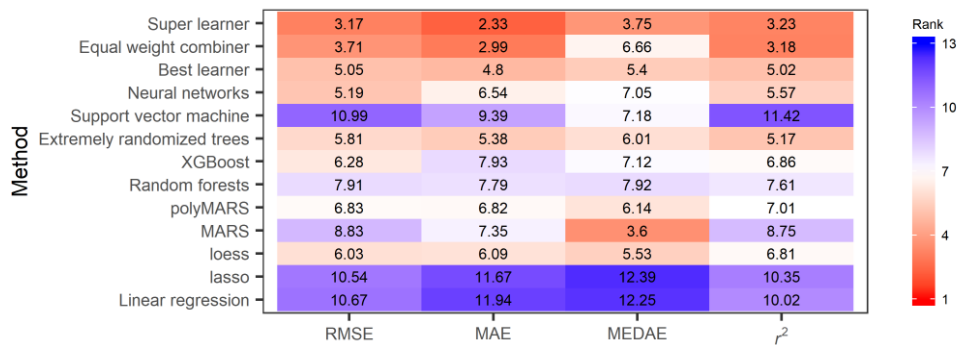

Figure 2. Mean rankings of the 13 algorithms according to their performance in the 511 basins.

Super learner is the best performing algorithm in terms of RMSE and MAE, and the second-best algorithm in terms of MEDAE and $r^2$; nonetheless, its difference from the best performing algorithm in terms of $r^2$ (i.e. the equal weight combiner) is minimal. In terms of RMSE, the equal weight combiner is the second-best performing algorithm, followed by the best learner. From the base-learners, neural networks, extremely randomized trees and loess are the best performing algorithms (ranked from best to worst) in terms of RMSE, while support vector machines are worse compared to the linear regression benchmark.

When focusing on metrics other than RMSE, one sees that the rankings of the algorithms remain similar, albeit not identical. For instance, while MARS is not well performing in terms of RMSE, it is the best performing learner in terms of MEDAE, contrary to the equal weight combiner, which does not perform well.

Figure 3 presents the rankings of the 13 algorithms according to their performance in terms of RMSE for the 511 basins considered. While, in general, one sees similar rankings of an algorithm at all basins (i.e. similar colours dominate a given row), there are cases where the rankings of an algorithm vary with respect to its mean performance. Take, for instance, the super learner. While it is on average the best performing algorithm, there are cases where other algorithms perform better.

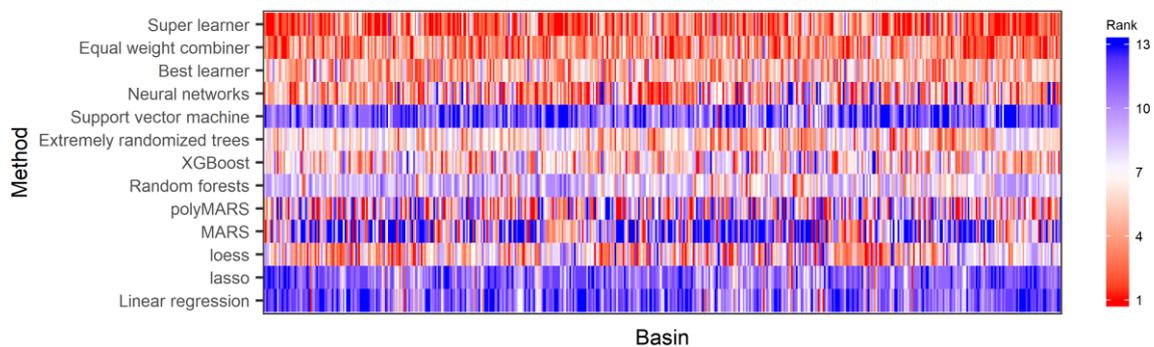

Figure 3. Rankings of the 13 algorithms according to their performance in terms of RMSE for the 511 basins considered.

### 4.2 Relative improvements

The mean relative improvement introduced by each algorithm with respect to the linear regression benchmark, is important for understanding whether a more flexible (yet less interpretable) algorithm is indeed worth implementing. In this context, Figure 4 presents the percentage of decrease of the RMSE, MAE, MEDAE and $r^2$ introduced by each of the 13 examined learners relative to that of the linear regression benchmark.

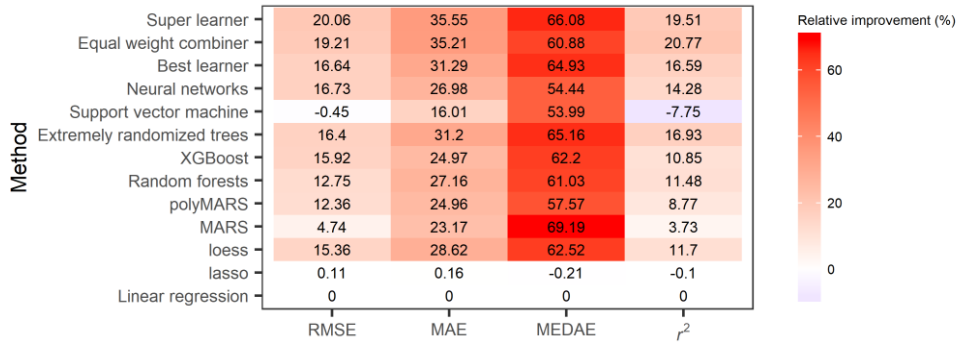

Figure 4. Mean relative improvements of the 13 algorithms with respect to the linear regression benchmark in the 511 basins considered.

Focusing on RMSE, super learner improves over the performance of the linear regression algorithm by 20.06%. The improvement introduced by the equal weight combiner is equal to 19.21% (not negligible as well), followed by the best learner with relative improvement equal to 16.64%. The best base-learner is neural networks, which improves over the performance of the linear regression algorithm by 16.73%, followed by extremely randomized trees (16.4%), XGBoost (15.92%), and loess (15.36%).

An important note to be made here is that the specific ranking of an algorithm in terms of the improvement it introduces relative to the linear regression benchmark, depends significantly on the metric used; i.e. RMSE, MAE, MEDAE and $r^2$. For instance, while the equal weight combiner is the second best-performing learner in terms of RMSE, MAE and $r^2$, it is the fourth worst performing in terms of MEDAE. In addition, please note that the magnitudes of the relative improvements differ considerably for the various metrics. For instance, relative improvements in terms of MAE mostly range between 25-35%, while the respective relative improvements in terms of RMSE are mostly between 10-20%.

To facilitate understanding of the range of forecast errors, Figure 5 presents boxplots of the RMSE values for all 13 algorithms considered. While in most cases the forecast errors lie below 5 mm/day, one sees that MARS and PolyMARS forecasts may fail considerably (see the exceptionally high outliers), and this is the case for neural networks as well. This form of instability could also explain why MARS and polyMARS are amongst the best performing methods in terms of MEDAE (a metric based on medians), while they appear to be less performing when assessed using metrics based on mean errors (i.e. RMSE, MAE and $r^2$).

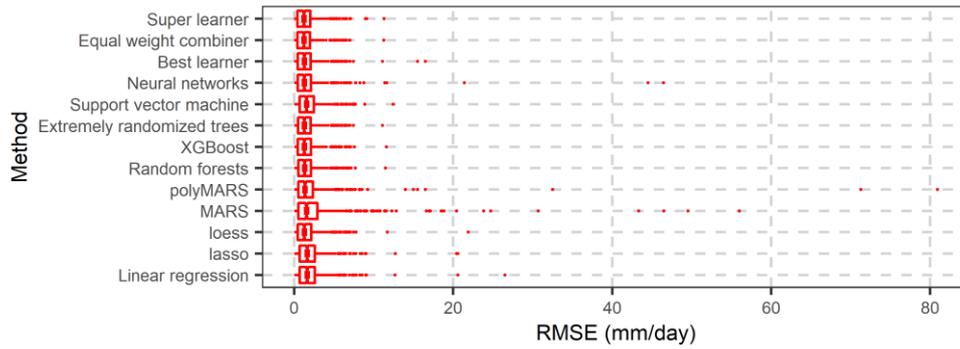

Figure 5. Boxplots of the RMSE values computed for the 13 algorithms in the 511 basins considered.

Values of $r^2$ are also of interest. Close inspection of Figure 6 reveals that the super learner and the equal weight combiner display values that lie mostly in the range 0.60-0.65, while the best learner exhibits somewhat lower values. The remainder base-learners display, in general, lower $r^2$ values, while the mean $r^2$ of linear regression is somewhat higher than 0.5.

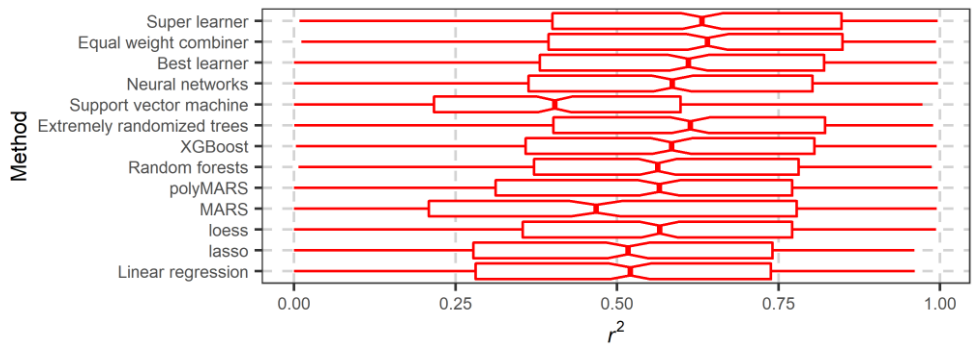

Figure 6. Boxplot of the $r^2$ values computed for the 13 algorithms in the 511 basins considered.

4.3 Weights

The weights of the base-learners (used to compose the super learner) are strongly linked to the performance of the 10 base-learners in the test set. This is becomes apparent from Figure 7, which presents the weights assigned to the 10 base-learners per basin. More precisely, close inspection of Figure 7 alongside with Figure 4, reveals that less performing algorithms in the test set are assigned smaller weights.

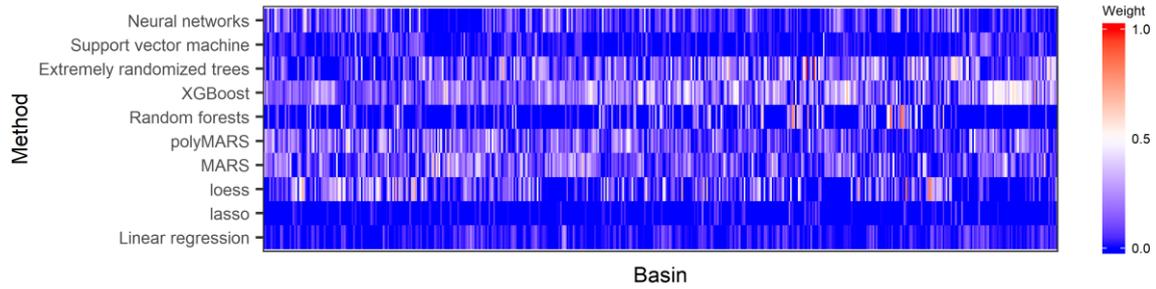

Figure 7. Weights assigned to the 10 base-learners in the 511 basins considered.

The boxplots in Figure 8 also confirm the aforementioned observation/finding; i.e. that the best performing methods in the cross-validation set (i.e. methods that are assigned higher weights) are those displaying the highest performance in the test set (see also Figure 4). The highest weights are assigned to XGBoost, which is one of the best performing algorithms.

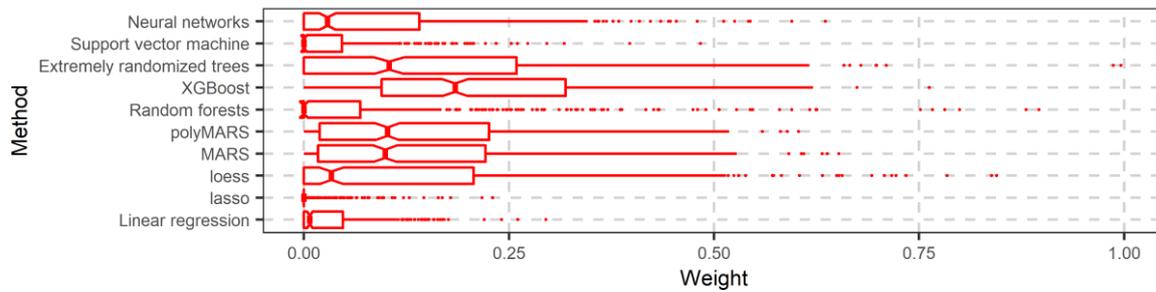

Figure 8. Boxplot of the weights assigned to the 10 base-learners in the 511 basins considered.

The boxplots in Figure 9 summarize results from all basins considered, and present how the weights assigned to the base-learners composing the super learner, are related to their individual rankings within the testing period. Clearly, the higher the weight, the better the performance of the algorithm in the testing period.

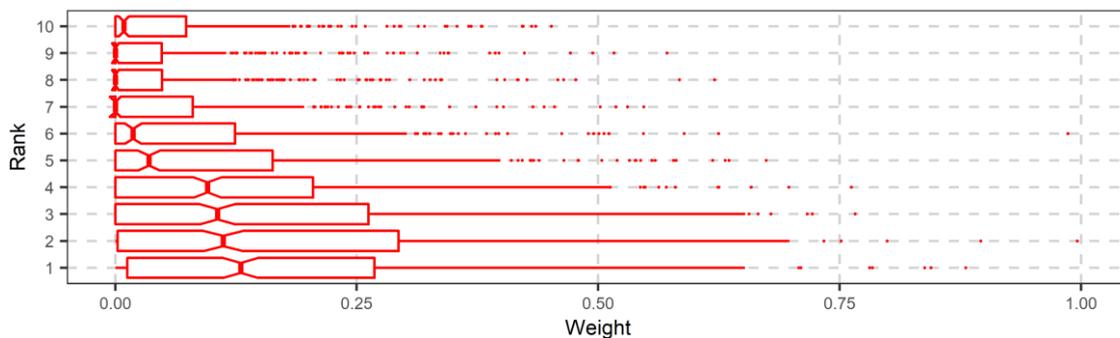

Figure 9. Boxplots of the weights assigned to the 10 base-learners conditional on their ranking in terms of RMSE.

## 5. Take-home remarks

We presented a new method for daily streamflow forecasting. This method is based on super learning. The introduced algorithm combines 10 base-learners and was compared to an equal weight combiner and a best learner (identified in the cross-validation procedure). We applied the algorithms to a dataset consisting of 511 river basins with 10 years of daily streamflow, precipitation and temperature. The machine learning algorithms modelled the relationship between next-day streamflow and daily streamflow, precipitation and temperature up to the present day.

The super learner improved over the performance of the linear regression benchmark by 20.06% in terms of the RMSE, while the respective improvements provided by the other ensemble learners were 19.21% (equal weight combiner) and 16.64% (best learner). The best base-learner was neural networks (16.73%), followed by extremely randomized trees (16.40%), XGBoost (15.92%), loess (15.36%), random forests (12.75%), polyMARS (12.36%), MARS (4.74%), lasso (0.11%) and support vector machines (−0.45%).

All ensemble learners improved over the performance of the single base-learners. The performance of the super learner was somewhat higher than that of the equal weight combiner, which according to the "forecast combination puzzle", is a "*hard to beat in practice*" combination method. Consequently, we consider that the equal weight combiner can be effectively used as a benchmark for new combination methods, while super learning can result in better performances. One could claim that based on statistical tests this difference may be insignificant, however as mentioned by [6] these tests should be avoided when comparing forecasting methods, as they can be misleading.

An advantage of super learning is that it can be optimized with respect to any loss function; in our case this loss function was RMSE. Albeit base-learners may be designed to optimize other loss functions, a combination approach (such as the super learner proposed herein) may be useful to extract their advantages with respect to a specific loss function. In general, other loss functions could also be used for optimizing the super learner.

We emphasize that our results are based on a big dataset; therefore, the reported relative improvements against the benchmark (in the range 0-20% in terms of RMSE) can be considered realistic and can provide insights and serve as a guide to understand whether results reported in the literature (single case studies indicating improvements more than 50%) could be attributed to chance related to the use of small datasets.

Regarding the usefulness of the proposed method one should consider that it is fully automated and does not need any assumptions, since it exploits a *k*-fold cross-validation procedure (in contrast e.g. to Bayesian model averaging, which is widely used in hydrology). Future research could focus on improving the variable selection procedure and comparing the ensemble learner with optimized base-learners, while testing on different datasets could be also useful.

**Conflicts of interest:** We declare no conflict of interest.

**Appendix A    Used software**

We used the `R` programming language [55] to implement the algorithms of the study, and to report and visualize the results.

For data processing, we used the contributed `R` packages `data.table` [22], `gdata` [81], `readr` [85], `stringi` [30], `stringr` [83], `tidyr` [84].

The algorithms were implemented by using the contributed `R` packages `earth` [44], `extraTrees` [61,62], `glmnet` [28,29], `kernlab` [37,38], `Matrix` [8], `nnet` [58,79], `polspline` [39], `ranger` [90,91], `SuperLearner` [54], `xgboost` [18].

The performance metrics were computed by implementing the contributed `R` package `mlr` [9,10].

Visualizations were made by using the contributed `R` package `ggplot2` [82,86].

Reports were produced by using the contributed `R` packages `devtools` [87], `knitr` [92,93,94], `rmarkdown` [5,95].